\documentclass[final]{cvpr}

\usepackage{times}
\usepackage{epsfig}
\usepackage{graphicx}
\usepackage{amsmath}
\usepackage{amssymb}
\usepackage{multirow}
\usepackage[table,xcdraw]{xcolor}

\newcommand{\sign}{\text{sgn}}

\usepackage[pagebackref=true,breaklinks=true,colorlinks,bookmarks=false]{hyperref}

\newcommand{\thickhline}{%
    \noalign {\ifnum 0=`}\fi \hrule height 1pt
    \futurelet \reserved@a \@xhline
}

\DeclareRobustCommand\onedot{\futurelet\@let@token\@onedot}
\def\onedot{\ifx\@let@token.\else.\null\fi\xspace}

\def\etal{\emph{et al.}}

\def\cvprPaperID{11146} % *** Enter the CVPR Paper ID here
\def\confYear{CVPR 2021}

\begin{document}

%%%%%%%%% TITLE
\title{LEAD: LiDAR Extender for Autonomous Driving}

\author{Jianing Zhang\\
Tsinghua University\\
{\tt\small jn-zhang19@mails.tsinghua.edu.cn}
\and
Wei Li\\
Inceptio\\
{\tt\small wei.li@inceptio.ai}

\and
Honggang Gou\\
Inceptio\\
{\tt\small honggang.gou@inceptio.ai} 

\and
Lu Fang\\
Tsinghua University\\
{\tt\small fanglu@tsinghua.edu.cn}

\and
Ruigang Yang\\
Inceptio\\
{\tt\small ruigang.yang@inceptio.ai}

}

\maketitle

%%%%%%%%% ABSTRACT
\begin{abstract}
   3D perception using sensors under vehicle industrial standard is the rigid demand in autonomous driving. MEMS LiDAR emerges with  irresistible trend due to its lower cost, more robust and meeting the mass production standards. However, it suffers small field of view (FoV), slowing down the step of its population. In this paper, we propose LEAD, i.e., LiDAR Extender for Autonomous Driving, to extend the MEMS LiDAR by coupled image w.r.t both FoV and range. We propose a multi-stage propagation strategy based on depth distributions and uncertainty map, which shows effective propagation ability. Moreover, our depth outpainting/propagation network follows a teacher-student training fashion, which transfers depth estimation ability to depth completion network without any scale error passed. To validate the LiDAR extension quality, we utilize a high-precise laser scanner to generate ground-truth dataset. Quantitative and qualitative evaluations show that our scheme outperforms SOTAs with a large margin. We believe the proposed LEAD along with the dataset would benefit the community w.r.t depth researches.
\end{abstract}

\vspace{-10pt}
%%%%%%%%% BODY TEXT

\section{Introduction}

Autonomous driving (AD) is one of the most challenging problems in computer vision and artificial intelligence, which has attracted considerable attention in recent years. With unremitting great efforts from academia and industry, the landing of AD system is around the corner. Among all of the advances AD system should achieve for mass production, 3D perception using sensors under vehicle industrial standard is the rigid demand in the near future. To meet this goal, more and more types of MEMS LiDAR emerge, with lower cost, more robust performance and most importantly, meeting the mass production standards comparing with traditional mechanical LiDAR sensors.
However, researches on 3D perception, including but not limited to depth estimation, 3D detection and 3D semantic segmentation, still focus on data from mechanical LiDAR sensor. In this paper, we tackle the problem of depth completion/estimation based on the MEMS LiDAR. 

%The dense depth map which is essential in environment perception, target recognition, path planning, and so on, plays a vital role in autonomous driving. There are some ways to generate dense depth map, stereo matching~\cite{chang2018pyramid}~\cite{zhang2019ga}, depth completion~\cite{uhrig2017sparsity}~\cite{ma2019self}, monocular depth estimation~\cite{godard2017unsupervised}~\cite{godard2019digging}, however, stereo matching requires precise calibration and robust mechanical structure, depth completion relies on expensive mechanical LiDAR which also limited the depth detection range and monocular depth estimation is an ill-posed problem affected by scale ambiguity. Considering these problems, we propose a new depth estimation method based on a new type of LiDAR, solid-state LiDAR, a.k.a MEMS LiDAR.

\begin{figure}
    \centering
    \includegraphics[width=\linewidth]{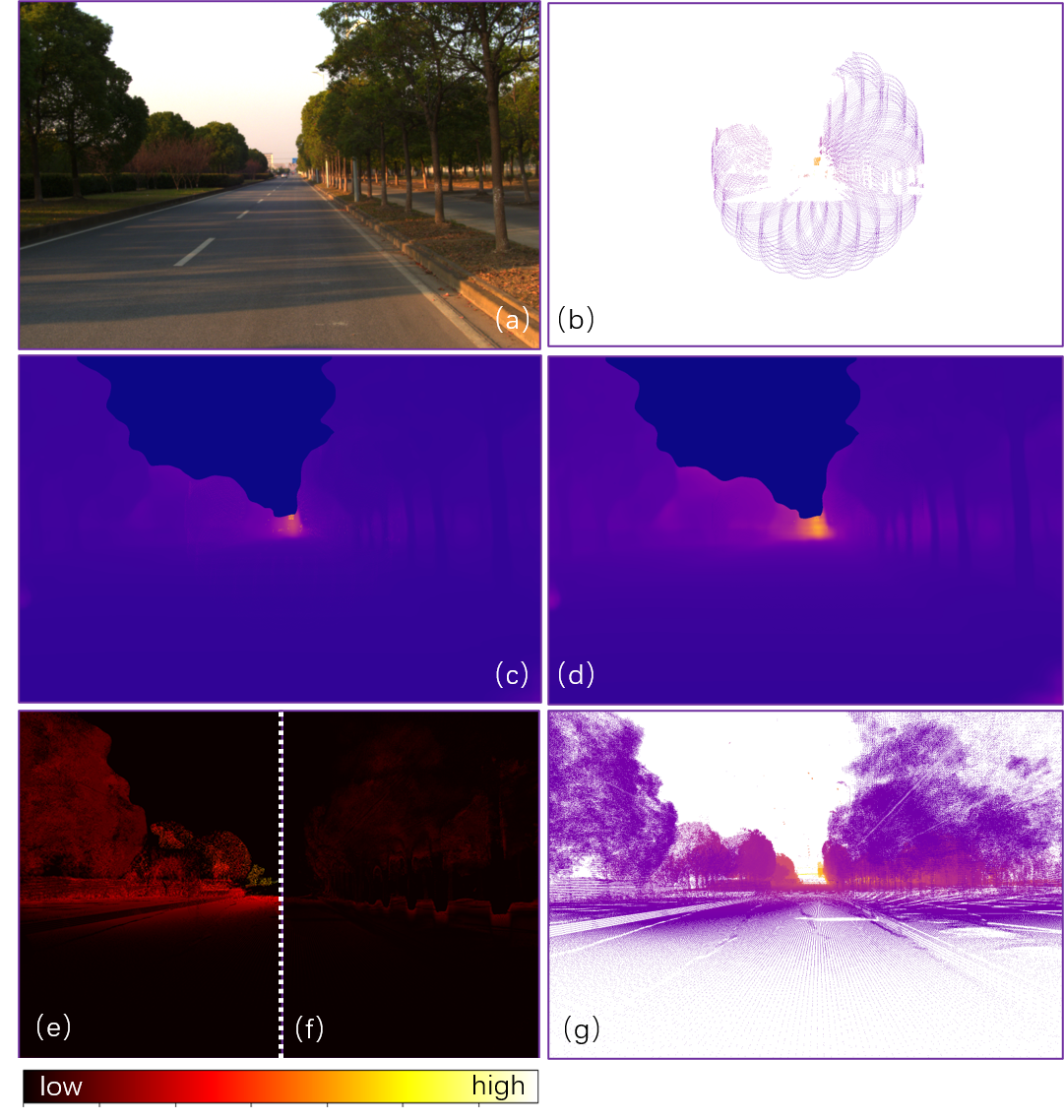}
    % \vspace{-12pt}
    \caption{(a) and (b) are the input pair, i.e., reference image from RGB camera and partial depth map from MEMS LiDAR. (c) is the improved result of MonoDepth2~\cite{godard2019digging}. (d) is our depth extension result. (e) and (f) are the error map of MonoDepth2~\cite{godard2019digging} and ours respectively. (g) is the ground truth from our LEAD dataset.} 
    %Extend the small FoV depth map. (a) is the reference image, (c) is the input small FoV depth map generated by MEMS lidar, (b) is the baseline result, (d) is our result, (e) is the ground truth, (f) is the error map of baseline method ,and (g) is the error map of our method.}% The buildings and goal in our final depth result keeps decent continuity and integrity}
    \vspace{-12pt}
    \label{fig:teaser}
\end{figure}
Despite the irresistible trend of using MEMS LiDAR, it suffers small field of view (FoV), which slows down the step of its population. Typical mechanical LiDAR sensors are always with $360^\circ$ FoV, while MEMS LiDAR such as the one adopted in our paper is just $14.5^\circ \times 16.2^\circ$. A promising direction is utilizing a coupled camera with larger FoV to extend the point cloud/depth from MEMS LiDAR. A natural question to ask is: why do not we directly use cameras for depth estimation if an extra camera is introduced? It is true that depth estimation from monocular or stereo cameras, especially with the power of deep learning, achieves compelling results~\cite{eigen2014depth,godard2017unsupervised,zhou2017unsupervised,godard2019digging,garg2016unsupervised,poggi2020uncertainty}. For example, some works ~\cite{garg2016unsupervised,godard2017unsupervised} utilize left-right consistency cues to train the monocular depth estimation network. Some online estimate poses from sequential images, and then form a photometric re-projection loss in sequence for training. Nevertheless, the depth maps estimated in this line of works suffer from scale ambiguities. They may yield plausible relative depth map, but hard to deploy in real AD systems due to the incredible scale. 

Another family of works avoid the scale ambiguities with the guidance of LiDAR even only partial observations w.r.t the FoV of camera. The problem then becomes how to complete the depth maps with images~\cite{uhrig2017sparsity,qiu2019deeplidar}. Regarding existing depth completion approaches, the input depth maps are projected from raw point clouds from mechanical LiDAR. Thus, the depth pattern is sparse, but covers the whole space of target view (the FoV of image). Those methods focus on \emph{inpainting} holes of depth map using interpolation and propagation mechanism~\cite{cheng2019learning,park2020non}. Oppositely, as shown in~\figref{fig:teaser}, our MEMS LiDAR could only produce very limit observations in a small area, which usually lies on the center of coupled images. In other words, we need to \emph{extend/outpainting} significantly from partial depth to fill the whole space of target image. Actually, there are few works of depth completion on line sensors~\cite{liao2017parse}, and they are only validated on indoor scenes. To the best of our knowledge, our approach is the first one of extending such limited depth observation to large FoV with high accuracy.

% The target of our LiDAR extender is to complete the input depth map, generate a large depth map based on the input depth map, and extend the depth detection range to the distant area. Considering that collecting the distant dense depth ground truth is tedious, we adopt a self-supervised learning framework that only requires input sequences for training, and for our special task, we propose depth propagation and distribution output module based on GAN. Our method can also generate the uncertainty map and support different extension modes. In addition to the software, we also construct our hardware system and introduce our dataset for MEMS LiDAR. Our contributions are summarized as follows: 

The key insight of the proposed approach is to propagate partial depth from MEMS LiDAR to a larger area with the guidance of image. We experimentally found directly propagating such small area of depth is impractical. Thus, we introduce a multi-stage propagation to deal with this problem gradually. In each stage, we extend current high confident area, which is the depth from MEMS LiDAR at the first stage, outward with a certain width. Eventually, we can get full of accurate depth after several stages. However, an effective extending operation between two stages is not trivial. Technically, in our approach, the output of one stage are a depth distribution set and an uncertainty map instead of one simple depth map. Then we use a novel probabilistic selection and combination operator to yield the depth for next state. With these mechanism, accurate depth could effectively propagate out. However, the network stays fragility even with well-designed training strategy or parameters. Thus, we propose a teacher-student training fashion to learn depth estimation ability from existing monocular depth estimation algorithm, while not mislead by its wrong scale. We build a teacher network based on Mono2\cite{godard2019digging} for our core depth outpainting/propagation network. Qualitative and quantitative experiments show that the proposed method outperforms SOTAs with a large margin. Additionally, in order to have accurate metrics, we use a 3D high-accurate long range laser scanner to collect our own dataset. In summary, the technical contributions of our work are as follows:

% As shown in~\figref{fig:pipeline} our whole networks consist of the teacher network, student network, and discrimination networks. The teacher network can generate the initial depth map and distill the knowledge to teach the student network. Different from the teacher network, there are five output levels whose output area size increase gradually in our student network, and instead of predicting the depth maps, on each level, the student network predicts the depth distribution which can better propagate the partial depth step by step. Besides, to further correct the mapping relation between the RGB image and depth map, we regard the student network as the generator and introduce a set of discriminators for each output level. During the interface, without ground truth, we can generate a high-quality extended FoV depth map and corresponding uncertainty map. 

\begin{itemize} 
\item We introduce a new setup focusing on extending the MEMS LiDAR by coupled image w.r.t both FoV and range. We propose a multi-stage propagation strategy based on depth distributions and uncertainty map, which shows effective propagation ability.
\item We introduce a teacher-student strategy to combine monocular depth estimation and completion networks. We experimentally find such strategy could transfer depth estimation ability to depth completion network without any scale error passed. 
\item To validate the LiDAR extension quality, we utilize a high-precise laser scanner to generate ground-truth dataset. The dataset contains 16792 pairs of MEMS LiDAR data and image for training, as well as 120 pairs of panoramic depth with up to 5mm accuracy in 2500m range for testing. We hope such carefully collected data could benefit the community w.r.t depth researches.

% \item We propose a new self-supervised learning-based method to extend the depth map. We design our network structure according to the characteristics of the input data. We adopt the self-teaching network structure to refine the depth map, introduce the depth propagation module to propagate the small FoV depth map, and propose a distribution output module to correct the depth value distribution. High quality extended depth maps can be generated through our pipeline and we can also generate the uncertainty map to make our result uncertainty-aware.
% \item We 
\end{itemize}

% (1) New hardware. We introduce a new hardware system to generate a dense depth map and avoid the disadvantages of existing hardware systems. We focus on a new kind of LiDAR, which may replace the traditional LiDAR in the future and the related algorithm is urgently needed at present. 

% (2) New algorithm. We propose a new self-supervised learning-based method to extend the depth map. We design our network structure according to the characteristics of the input data. We adopt the self-teaching network structure to refine the depth map, introduce the depth propagation module to propagate the small FoV depth map, and propose a distribution output module to correct the depth value distribution. High quality extended depth maps can be generated through our pipeline and we can also generate the uncertainty map to make our result uncertainty-aware.

% (3)New dataset. The new hardware and new algorithm require the new dataset. Our new dataset is the first dataset that focuses on MEMS LiDAR, we are sure that our dataset will have a significant impact on the future development of this field.
%------------------------------------------------------------------------
\begin{figure*}[t]
	\centering
	\includegraphics[width=0.8\textwidth]{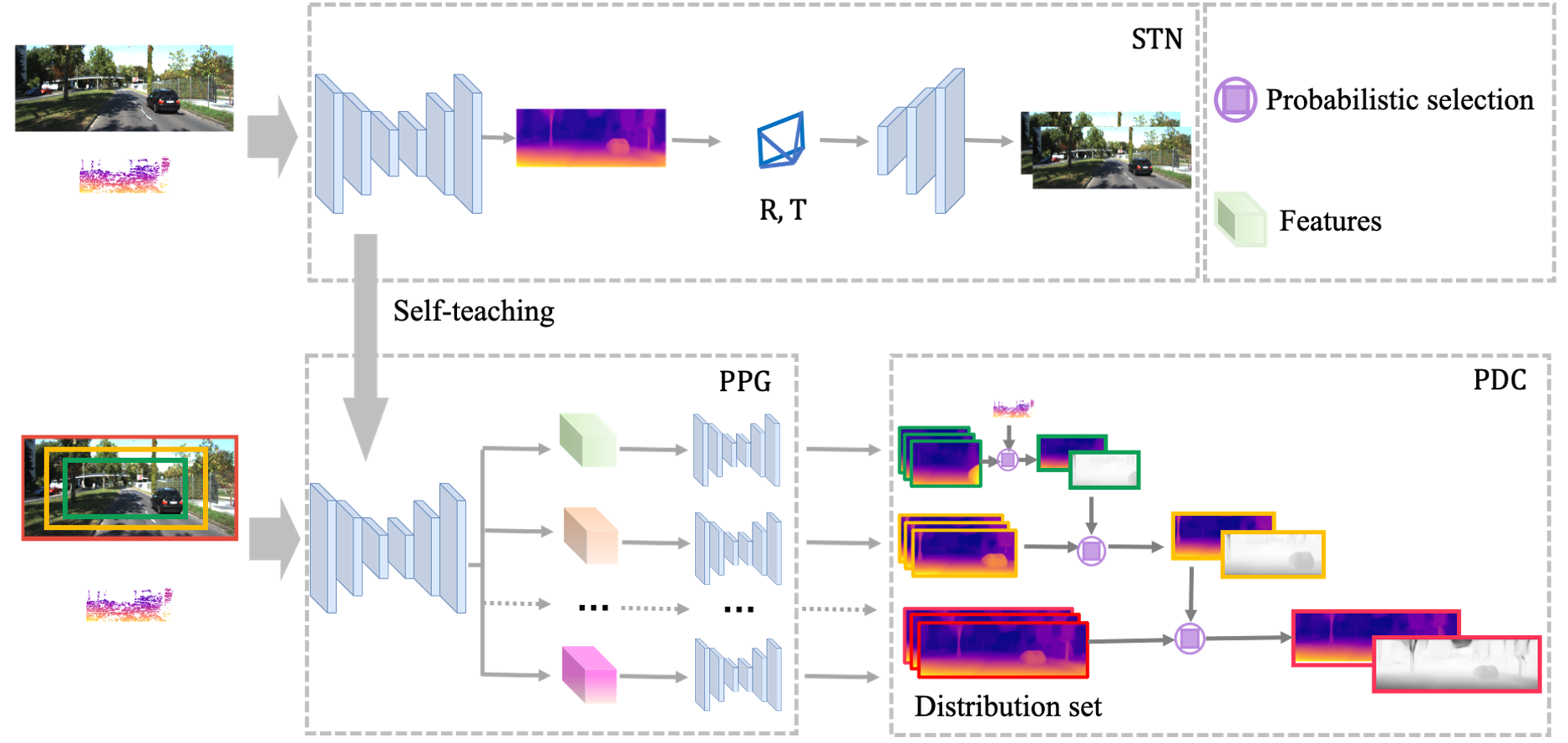}
% 	\vspace{-18pt}
	\caption{Overview of our method. The pipeline consists of a self-supervised teacher network~(STN),  propagative probabilistic generating~(PPG) module and probabilistic derivation and composition~(PDC) module. STN provides the initial depth and stabilizes the training processing of PPG. PPG generates the depth probabilistic distributions for different propagation levels. With these distributions and the input partial depth map, PDC derives the final depth map with corresponding uncertainty.}
% 	\vspace{-12pt}
    \vspace{-12pt}
	\label{fig:pipeline}
\end{figure*}

\section{Related Work}

\subsection{Depth Completion}

% \cite{pinggera2014know, ma2019self, qiu2019deeplidar,mendes2020deep, senushkin2020decoder, xia2020generating}

Depth completion is a high-relevant topic with our task, which focuses on yielding dense depth from sparse or noisy point cloud data~\cite{qiu2019deeplidar, mendes2020deep, senushkin2020decoder}. The classical depth completion methods take only sparse or noisy depth sample from LiDAR or SLAM/SfM systems as input, which fall into the concept of depth super-resolution~\cite{lu2015sparse}, depth inpainting~\cite{zhang2018probability}, depth denoising~\cite{shen2013layer}. In this category, Zhang~\etal~\cite{zhang2018deep} proposed to predict the surface normal to estimate the dense depth map for indoor scenes in NYUv2 dataset\cite{silberman2012indoor}. Ma~\etal~\cite{ma2019self} use sparse depth maps and images to first calculate the camera poses, and then train a depth completion network based on predicted poses. Those works achieve compelling results of depth densification, but they only deal with partial depth map in the same resolution with RGB image, which is sparse samples in the complete spatial space. So, the network only needs to inpaint the empty part of the depth map. However, in our setup, the input depth is only small part w.r.t the vertical and horizontal space of our final estimated depth. Our task is more a depth outpainting problem with both the FoV and range extension are required. Instead of mechanical LiDAR, Liao~\etal~\cite{liao2017parse} adapt a line sensor to get the partial depth measurement and generate a dense depth map but limited by the setup, this method can only be applied indoor.

In depth completion, propagating high-confidential depth samples to the whole depth map is a widely adopt mechanism. Spatial Propagation Networks (SPN)~\cite{liu2017learning} trained a neural network to learn the local linear spatial propagation which can enhance the depth map or segmentation result but each time SPN updates a pixel, the whole image needs to be scanned, which affects the efficiency of the algorithm. Cheng~\etal~\cite{cheng2019learning} proposed CSPN which focuses on the local context to propagate the depth. Since the depth value is generally closely related to the surrounding depth values, the local propagation can improve the depth map efficiently. The CSPN based method further improved efficiency by adding resource-aware module~\cite{article} and non-local information~\cite{park2020non}. The depth propagation method also has the potential to extend the small FoV depth, ~\cite{cheng2020ode} propagate the small FoV depth to generate the panoramic depth.

\subsection{Depth Estimation}

The target of monocular depth estimation is to estimate the depth map with a single camera. This problem is ill-posed since an image can project to many plausible depths. So in practice, these methods can only provide an estimation or relative depth map. Eigen~\etal ~\cite{eigen2014depth} proposed the first learning-based monocular depth estimation algorithm relies on the dense ground truth. After this work, a lot of supervised learning based monocular depth estimation merged, such as ~\cite{fu2018deep} and ~\cite{guo2018learning}. However, acquisition of the accurate and dense ground truth is tedious. ~\cite{mayer2018makes} tried to explore the potential of synthetic training data but the complexity of synthetic data is still not comparable with the real data.
To break through the limitation of insufficient training data, Gard~\etal~\cite{garg2016unsupervised} proposed a self-supervised learning framework that uses the stereo pair and constructs the stereo photometric reprojection warping loss to train the network. After that, Godard~\etal~\cite{godard2017unsupervised} improved stereo based framework by adding a left-right consistency loss. In addition to stereo pair, Zhou~\etal~\cite{zhou2017unsupervised} only used the video sequences to estimate both the camera pose and depth map and the assumption of this method is that the scene is static so the network had to predict a mask to filter out the moving objects. Godard~\etal~\cite{godard2019digging} also improved sequence-based self-supervised framework with an auto-mask, besides, in ~\cite{godard2019digging}, the two kinds of the self-supervised framework were used together, which also improved the monocular depth estimation result. To improve the quality of the depth map, some other tasks can be added and trained together, such as optical flow estimation~\cite{yin2018geonet} and image segmentation~\cite{zhu2020edge}~\cite{wang2020sdc}. To find the uncertain area, some algorithms~\cite{Eldesokey_2020_CVPR}~\cite{johnston2020self}~\cite{poggi2020uncertainty} generate the uncertainty map while estimating the depth map.

However, these monocular depth estimation methods are limited by the scale ambiguity. The supervised method can only recover the scale of the dataset, the scale of the stereo based self-supervised method is up to the baseline of the training dataset and the sequence-based method needs ground truth to correct the scale, which greatly reduces the practicability of the algorithm.

\begin{figure*}[t]
	\centering
	\includegraphics[width=0.8\textwidth]{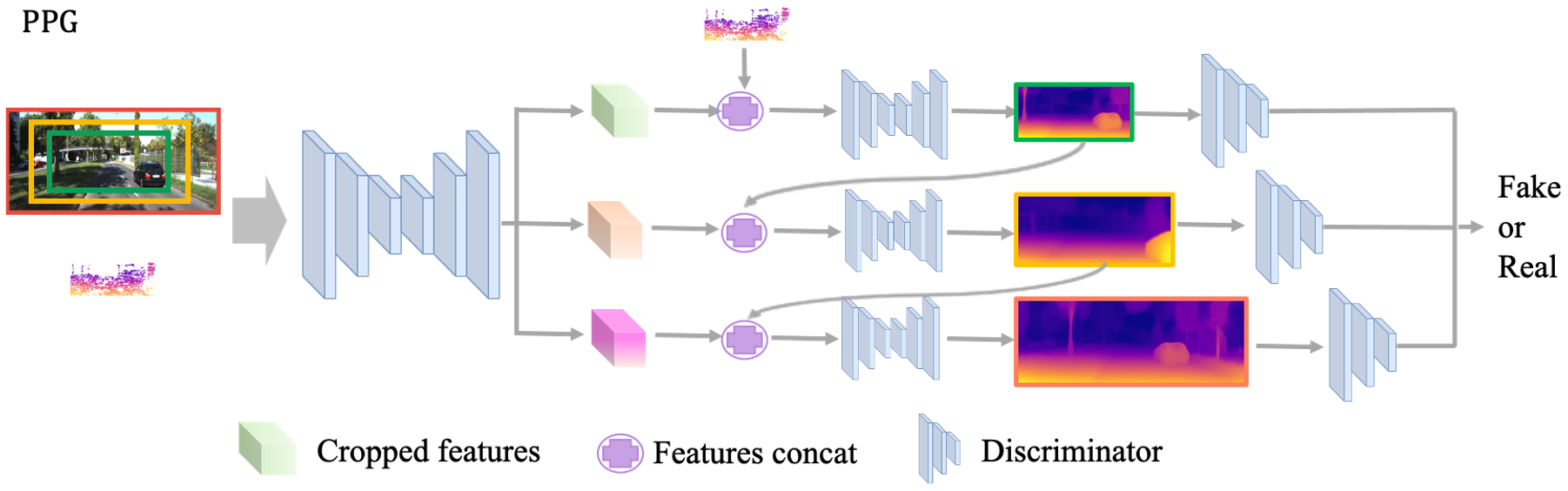}
% 	\vspace{-12pt}
	\caption{Propagative probabilistic generating module. For each propagation stage, the network can fuse the output depth from last stage to generate depth map for next stage. On each stage, there is a discriminator to judge the quality of generated depth map.}
 	\vspace{-12pt}
	\label{fig:PPG}
\end{figure*}

\section{Method}
\subsection{Overview}

%Here, we describe our depth prediction network that extends a small FoV depth map $D_{part}(t)$ to a large FoV dense depth map $D(t)$ with the guidance of the RGB image $I(t)$ without depth ground truth. 
Given the RGB image $I$ and partial depth map $D_s$ generated from MEMS LiDAR sensor, our goal is to estimate full depth map $D_{f}$ with same FoV of camera, in other words, the same resolution of image $I$. The key insight of the proposed method is to propagate the partial depth to larger areas with guidance of image.  Digging into the inputs, the partial depth $D_s$ provide the some accurate observations, and the structure in RGB image $I$ encodes plentiful cues for depth distribution so we want to learn a mapping $G: (I,D_s) \rightarrow p$. Due to the scale ambiguity, the mapping relationship is not unique and inspiring by~\cite{xia2020generating}, instead of estimating $D_{f}$ directly, our method estimate conditional depth distribution $p(D|I,D_s)$. With the depth distribution and small FoV depth map, we can derive the depth map with right scale. In our approach, we adopt GAN to generate a set of depth maps $\{d_1,d_2, \cdot\cdot\cdot,d_k\}$ to approximate depth distribution discretely.  

However, training such probabilistic generator faces some challenges. First, generating dense outdoor depth maps needs to face the challenge that acquiring the ground truth is hard. Even using the expensive lidar, there are still holes in the depth maps, so here we adopt the unsupervised training method. Second, training process is not stable, especially without ground truth. Besides, how to make full use of the partial depth and propagate the partial depth is also a problem.

To solve these challenges and extend the small FoV depth well, we propose self-supervised teacher network, probabilistic propagation generator and probabilistic derivation and composition module.

%As shown in~\figref{fig:pipeline} our whole networks consist of the teacher network, student network, and discrimination networks. The teacher network can generate the initial depth map and distill the knowledge to teach the student network. Different from the teacher network, there are five output levels whose output area size increase gradually in our student network, and instead of predicting the depth maps, on each level, the student network predicts the depth distribution which can better propagate the partial depth step by step. Besides, to further correct the mapping relation between the RGB image and depth map, we regard the student network as the generator and introduce a set of discriminators for each output level. During the inference, without ground truth, we can generate a high-quality extended FoV depth map and its corresponding uncertainty map. 

% \subsection{GAN Module and distribution output}

\subsection{Self-supervised Teacher Network (STN)}
To stabilize the training process without ground truth, we first train a teacher network to guide the probabilistic generator. The teacher network can provide the initial depth map for next stage, and the depth map can be used as the pseudo ground truth. Besides, the probabilistic generator that shares the encoder and decoder structure and initialize with the weight of teacher network can distill the knowledge learned by the teacher network. The input of our teacher net is RGB and partial depth map, however, since the effective region of partial depth map is small, if we directly concatenate the RGB and the partial depth map, the value distribution of final depth map is uneven and not smooth, to avoid this problem, we first extract features of the RGB and the partial depth map separately and concatenate the featuresa as the input of Unet. 

Without ground truth, we adopt the self-supervised training method. In addition to the network for depth estimation, we construct a network for pose estimation. With the estimated pose and depth we can synthesis the target image from another viewpoint. By minimizing the photometric reprojection loss, the network can be trained to generate depth maps. The estimated pose can also be used to supervised the probabilistic generator.
% \paragraph{Self-supervised depth estimation}

% % Generating dense outdoor depth maps needs to face the challenge that acquiring the ground truth is hard. Even using the expensive lidar, there are still holes in the depth maps, so here we adopt the self-supervised training method. In addition to the network for depth estimation, we construct a network for pose estimation. With the estimated pose and depth we can synthesis the target image from another viewpoint. By minimizing the photometric reprojection error, the network can be trained to generate depth maps.

% In our network, we concatenate RGB images with our small FoV depth maps as input to extract features and generate depth maps. Compared with the self-supervised monocular depth estimation, we can avoid the ambiguous scale problem by supervising results with our partial depth. However, in this way, the output depth maps are not accurate since we do not make full use of partial depth maps.

% \subsection{Teaching and Propagation Network}
% \subsection{Self-teaching learning and Depth Propagation}
% \paragraph{Self-teaching leanring}
% To improve accuracy, we introduce the self-teaching learning method. The depth network in the last section has already learned how to generate depth maps from inputs, we regard this network as the teacher network. Based on the teacher network, we build the student network which has the same encoder and decoder structure as the teacher network, and depth maps generated by the teacher network can be used as the pseudo ground truth to supervise the student network.
\subsection{Propagative Probabilistic Generator (PPG)}
Based on the guidance of STN, we can build our probabilistic generator. The target of the generator is to make full use of the partial depth map and generate the depth distribution. 
\paragraph{Multi stage propagation}
In our experiments, we found that the straightforward mapping of $I,D_s \rightarrow p$ is unsatisfactory. This is reasonable, as $D_s$ could provide relatively accurate depth values but only in very limited area, which cannot effectively propagate out to the whole space. Thus depth values in remaining area heavily rely on the inference from monocular image, which cannot guarantee the accuracy. Our insight here is a multi-stage propagation mechanism to correct distribution set incrementally in spatial space.
Considering the small FoV depth cannot affect the whole depth map but it can improve the nearby depth value, here we gradually expand the propagation area. For example, in the first propagation stage, we only fuse the small FoV depth with a little larger cropped depth map. During the fusion step, we pad the small FoV depth to the same size as the larger one and adjust the scale of the unrefined larger cropped depth maps based on the median ratio,
\begin{equation}
\begin{split}
    D_{i}^{s} = median(D_{i-1}>0)/median(D_{i}^{b}>0) \cdot D_{i}^{b},\\
    D_{i}^{m} = (1-\sign(D_{i-1}))D_{i}^{s}+\sign(D_{i-1})D_{i-1},
\end{split}
\end{equation}
where $D_{i}^{b}$ is the blur depth map cropped from the output of teacher network, $D_{i-1}$ is the refined depth map in stage $i-1$, in stage $i$, $D_{i-1}$ will be padded to the same size as $D_{i}^{b}$, $D_{i}^{s}$ is the depth after scale adjustment and $D_{i}^{m}$ is the mixed depth map. 
Then refine depth of this stage is generated with the concatenation of cropped RGB features and depth features extracted from the $D_{i}^{m}$. 
In the next stage, the last stage depth maps will be propagated into a larger cropped depth map. 
Repeating this step can propagate the small FoV depth into the whole depth map. For each stage, the generated depth map is based on the refined depth map of the last stage. 
\paragraph{Probabilistic Generating}

% In stage $i$, the uncertainty map is
% \begin{equation}
%     U_i=\sigma(p_i)=\sqrt{\frac{1}{N}\sum_{k=1}^N(d_k-\mu(p_i))},d_k\in p_i,
% \end{equation}
% where $U_i$ is the uncertainty map of stage $i$ and $N$ is the size of set $p_i$.

After propagation, to generate the depth distribution, we introduce a distribution generating block to generate the distributions of each propagation stage. ~\figref{fig:PPG} shows the details of our distribution generating block. While training, at each propagation stage, the generator can generate the depth map of this stage and the discriminator will judge the quality of the generated depth map. Like other generative adversarial networks, we introduce some noise to the input feature by randomly dropping out. While testing, we still enable the function of dropping out and generate probabilistic distribution by multiple forward sampling. Additionally, we can derive the uncertainty map $U$ using standard deviation of distribution set $p(D|I,D_s)$:
\begin{equation}
    U=\sigma(p)=\sqrt{\frac{1}{N}\sum_{k=1}^N(d_k-\mu(p))},d_k\in p,
\end{equation}
where $N$ is the size of distribution set $p$. $\mu(p) = \frac{1}{N}\sum_{k=1}^N d_k$ is the mean of $p$.

%map $I$ and $D_{p}$ into depth distribution
% Digging into the inputs, with the partial depth and corresponding RGB image, we can constrain the depth distribution according to the RGB image structure, in other words, the mapping $G:RGB \rightarrow D$ can be learned from the given images and small FoV depth map and be applied to the larger area to improve the result.

% To learn the mapping function, we adopt the generative adversarial network structure. The discriminator determine the RGBD input is true or not.

% With the student network, the partial depth has already been propagated into the whole depth maps. However, sometime after propagation, the partial depth value can be changed so during evaluating, we can use the partial depth to correct the scale. 

% Consider this task from another perspective, instead of predicting the depth map, given the partial ground truth we can generate the depth map by predicting the conditional distribution $p(D|I,D_{s})$, where $D$ is our output depth map, $I$ is the RGB image and $D_{s}$ is the partial depth map.

%Repeating this step can propagate the small FoV depth into the whole depth map. For each stage, the generated depth map is based on the refined depth map of the last stage. 

\begin{figure}
    \centering
    \includegraphics[width=\linewidth]{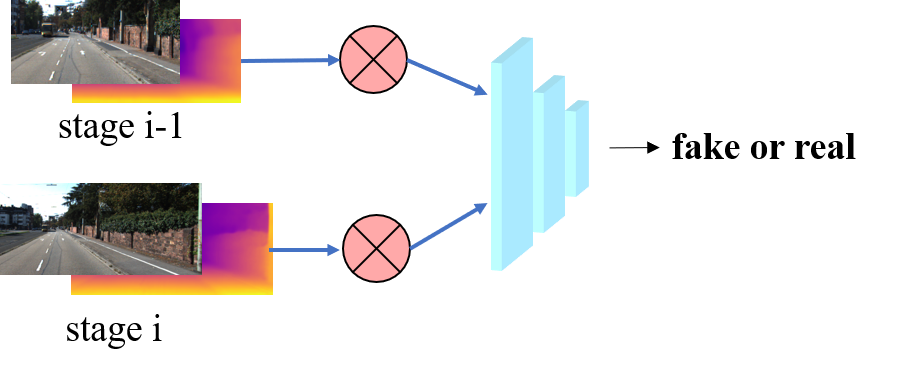}
    % \vspace{-12pt}
    \caption{The training process of the discriminator. The discriminator can learn the mapping relationship between RGB and D. At stage i, we regard the RGBD set of stage i-1 as ground truth and judge whether the RGBD set of i is fake or real.}% The buildings and goal in our final depth result keeps decent continuity and integrity}
    %\vspace{-12pt}
    \label{fig:gan}
\end{figure}

However, using the given small FoV RGBD values merely cannot train the discriminator well due to the limited field of view. We thus adopt the propagation strategy to train the GAN step by step. At each stage, the refined depth at previous stage is regarded as the true samples, while the cropped depth in this stage which aligns with RGB images of the previous stage is regarded as the false samples. In this way, the discriminator can gradually learn the right mapping function of RGB and D, and correct the depth map. Besides, for better training of the discriminator, we randomly change the scale of depth maps, i.e., in stage $i$, for the discriminator
\begin{equation}
\begin{split}
    D^{t}_{i} = S\cdot D_{i-1},
    D^{f}_{i} = S\cdot D_{i},
\end{split}
\end{equation}
where $S$ follows uniform distribution $S \sim U\text{[0.8,1.2]}$, we can change the size of the RGBD image by a random scale $S_{\text{size}} \sim U\text{[0.5,1.8]}$. The data augment forces the discriminator to learn the structural mapping relation between RGB and depth, while ignoring the different scales and sizes among different levels. In addition to the adversarial loss, we also use the pose and pseudo depth map learned by STN to train the generator.

%\paragraph{Multi-Stage Propagating}

\subsection{Probabilistic Derivation and Composition}

With the probabilistic depth distribution, the simple way to generate the final depth map is to calculate the mean of these distribution, however, this way cannot take advantage of partial depth map. ~\cite{xia2020generating} uses optimization algorithms to post process the network results, which increases the complexity of the algorithm. However, with our special propagative probabilistic generator, we can derive the final depth map without complex post-processing.
% Still, directly predict the $p(D|I,D_{s})$ is not easy, but we can predict the distribution $p(D_{i}|I_{i-1},D_{i-1})$ in stage $i$. So we adopt the GAN structure, in each stage, instead of generating only one depth map, we generate a set of depth maps to represent the depth distribution. The depth map in stage $i$ can be acquired by
Since we generate the depth distribution at each propagation stage, we can follow the same idea to derive and compose the final depth map stage by stage. The depth map $D_{i}$ in stage $i$ can be acquired by
\begin{equation}
\begin{split}
    D_{i} = \min_{d_{k,i}\in p_i}(\|(d_{k,i} - D_{i-1})\cdot \sign(D_{i-1})\| \\
    +\lambda\|(d_{k,i} - D_{s,i})\cdot \sign(D_{s,i})\|),
\end{split}
\end{equation}
where $p_i$ is the distribution set in $i$-th stage and in our experiment, at stage $i$, we run the generator for 5 times to generate the distribution set $p_i$, $D_{i-1}$ is the refined depth map in stage $i-1$, $d_k$ is the $k$-th depth map in $p_i$, $\lambda$ is the weight coefficient and $D_{s,i}$ is the cropped partial depth map whose size is the same as $D_i$. So in each stage, we make sure that the refined depth map is similar to the previous stage refined depth map and the partial ground truth. With $D_{i}$, the distribution of the next stage can be generated by next generator and we use the same way to generate $D_{i+1}$. In this way, we can gradually propagate the partial depth information into the final result.

\subsection{Network Training}
For self-supervised learning, given the camera intrinsic and the relative pose between the two frames, we can use the depth map to synthesis the image of a novel view and calculate the photo-metric loss.
\begin{equation}
    \mathcal{L}_{pe}=F(\pi(I(t'),K,R|t,D,I(t))),
\end{equation}
\begin{equation}
    F(I',I)=\frac{\alpha}{2}(1-\text{SSIM}(I',I))+\|I'-I\|.
\end{equation}
For STN, we adopt multi-scale output and use three frames to train the network. While training with three frames,  
\begin{equation}
    \mathcal{L}_{pe}=\sum_{s=0}^{4}w_{i}\min_{i\in{-1,1}}F(I(t+i),I(t)).
\end{equation}
The partial depth maps can supervise the output depth maps,
\begin{equation}
    \mathcal{L}_{part}=\sum_{s=0}^{4}\|D_s-\sign(D_{part})\|.
\end{equation}
Besides, we use edge-aware smooth loss to improve the smoothness
\begin{equation}
    \mathcal{L}_{smooth}=|\partial_{x}D|e^{-|\partial_xI|}+|\partial_{y}D|e^{-|\partial_yI|}.
\end{equation}
The final loss function for STN is 
\begin{equation}
    \mathcal{L}_{t}=w_{pe}\mathcal{L}_{pe}+w_{p}\mathcal{L}_{part}+w_{s}\mathcal{L}_{smooth}.
\end{equation}
When training the PPG, different from the STN, the output of the network are five different size depth maps corresponding to five propagation stages. The PPG follows the GAN structure, the loss function contains a GAN loss and an appearance loss. The GAN loss is
\begin{equation}
    \mathcal{L}_{GAN}=\sum_{i=0}^{5}\mathbb{E}_{I_{i}^{t},D_{i}^{t}}[logD(I_{i}^{t},D_{i}^{t})]+\mathbb{E}_{z_{i}}[log(1-D(G(z_{i})))].
\end{equation}
For appearance loss, since the STN has generated $D_{blur}$, we adjust the scale of the depth map to supervise the PPG
\begin{equation}
    s=median(D_{part}>0)/median(D_t\cdot \sign(D_{part})),\\
\end{equation}
\begin{equation}
    D_{pseudo}=s\cdot D_{blur},
\end{equation}
\begin{equation}
    \mathcal{L}_{pseudo}=F(crop(D_{pseudo}),D_{s}),
\end{equation}
The partial depth can adjust the scale of $D_{blur}$ to generate $D_{pseudo}$. Considering that $D_{pseudo}$ is not sufficiently accurate, we expect PPG can learn the structure of the depth map from $D_{pseudo}$ instead of the value. Thus in $L_{pseudo}$, the weight of SSIM term is larger.
While training the generator, we need to adjust all the camera translations estimated by the posenet to construct the photo-metric loss,
\begin{equation}
    \hat{t}={t} \cdot{s},
\end{equation}
where $t$ is the origin pose and $\hat{t}$ is the adjusted pose. With $\hat{t}$, We construct the photo-metric loss of generator denoted as $\mathcal{L}_{peg}$, then the loss function of PPG is given by
\begin{equation}
    \mathcal{L}_{s}=w_{peg}\mathcal{L}_{peg}+w_{pse}\mathcal{L}_{pseudo}+w_{G}\mathcal{L}_{GAN}.
\end{equation}
For training, we use dropout as the random noise of the input. For testing, we still enable the dropout function work and generate the distribution sets.

%------------------------------------------------------------------------
\begin{figure}
    \centering
    \includegraphics[width=0.6\linewidth]{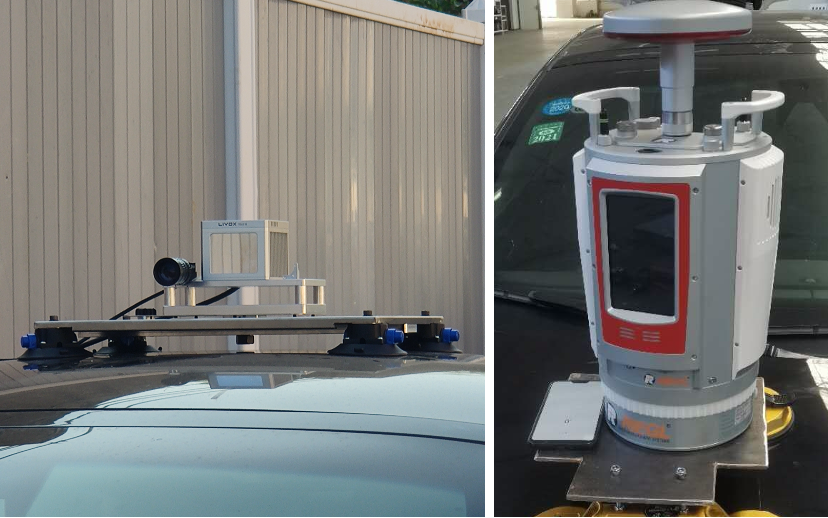}
    %\vspace{-15pt}
    \caption{\textbf{LEFT}: MEMS LiDAR and camera hardware setup. \textbf{RIGHT}: the laser scanner for ground truth depth estimation.}% The buildings and goal in our final depth result keeps decent continuity and integrity}
    % \vspace{-15pt}
    \label{fig:setup}
\end{figure}

\section{Experiment}
% In this section, we evaluate the performance of our method and compare it with other depth estimation method. We also introduce our ULRS dataset, which can provide accurate and distant ground truth.

\subsection{Hardware and Evaluation Dataset}

\paragraph{Hardware}
\figref{fig:setup} shows the MEMS LiDAR and camera hardware setup, which is mounted on the roof of a vehicle for outdoor scanning. 
We use the Tele-15 MEMS LiDAR sensor from Livox Inc.\footnote{https://www.livoxtech.com/tele-15} as the partial depth sensor. The FoV of the MEMS LiDAR is $14.5^\circ \times 16.2^\circ$. Although the claimed depth sensing range is 500 meters, the effective range is from 3 meters to 200 meters during our practical usage. In addition, we use an industrial camera from FLIR Inc. with 12mm focal length, $41.3^\circ \times 31.3^\circ$ FoV and $1536 \times 1024$ resolution. The MEMS LiDAR and camera are calibrated.

\begin{figure}
    \centering
    \includegraphics[width=\linewidth]{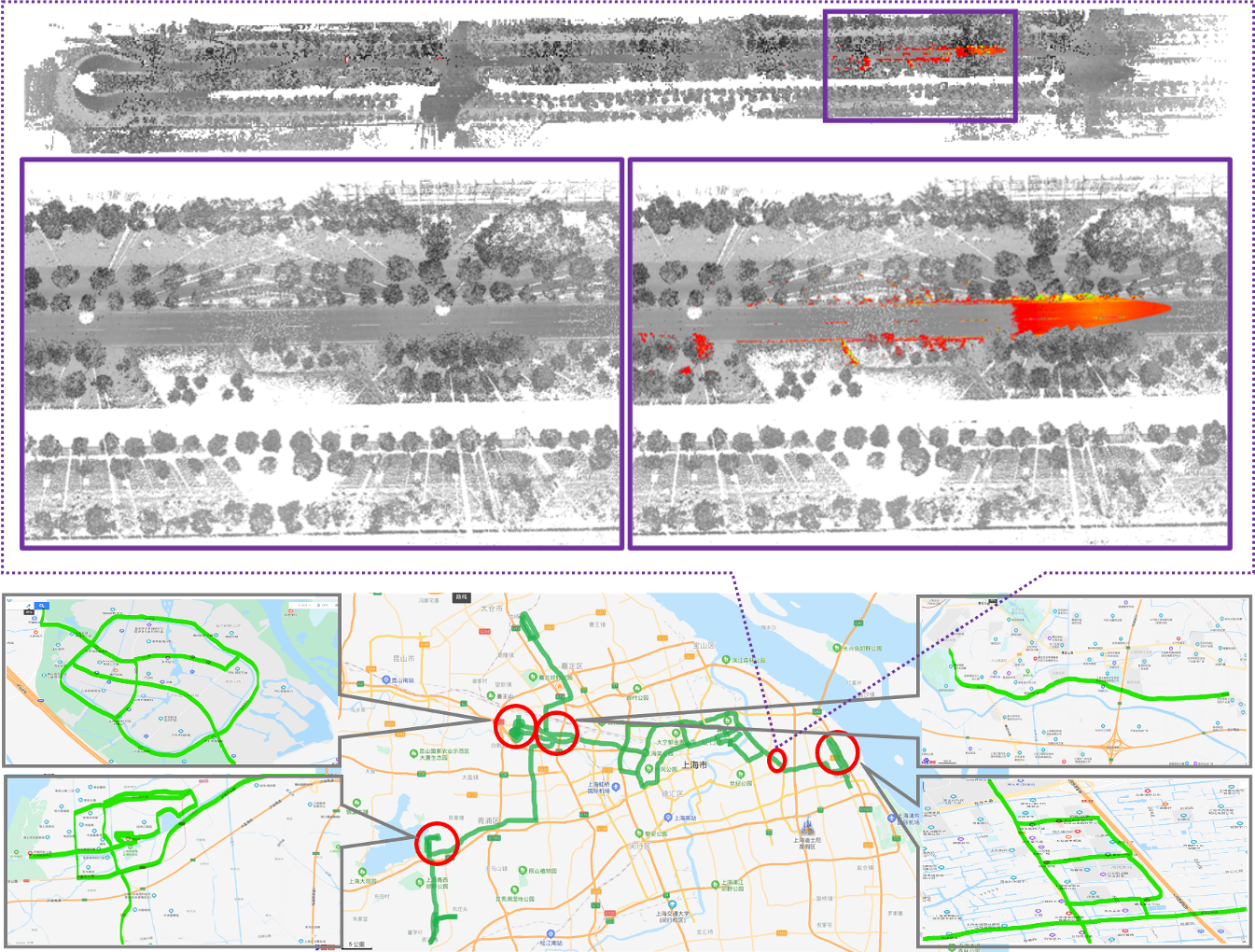}
    %\vspace{-12pt}
    \caption{Collection area and samples in LEAD dataset. The map in bottom row shows the data collection area. The point clouds in dotted box are captured from MEMS LiDAR (red) and high-accurate laser scanner (grey). }% The buildings and goal in our final depth result keeps decent continuity and integrity}
    % \vspace{-12pt}
    \label{fig:dataset}
\end{figure}
\paragraph{Our LEAD Dataset}

We captured around 100 hours data, in 50 streets over 500 kilometer from urban area in Shanghai, China. Then, we semi-automatically pick 16792 pairs of partial depth maps and their corresponding images, which are used as training and validation set.
To quantitatively evaluate the extended depth map from our setup, we utilize an additional high-accurate long range laser scanner to capture additional 120 pairs with ground truth depth. We use Riegl Inc.'s vz-2000~\footnote{http://www.riegl.com/nc/products/terrestrial-scanning/produktdetail/product/scanner/58/} 3D terrestrial laser scanner. The scanner has a wide FoV of $100^\circ \times 360^\circ$ with up to 2500 meters sensing capability and 5 mm accuracy. We calibrate the laser scanner and the MEMS LiDAR by align the dense depth and the partial depth using point-to-plane ICP. \figref{fig:dataset} shows one sample of testing set in our LEAD dataset. We are releasing the full LEAD dataset to public along with this paper. We believe it will benefit the community including but not limited to the research on MEMS LiDAR depth estimation.

%Our target is to extend the depth map generated by the MEMS LiDAR with a relatively large field of view camera so we construct our hardware system and propose the corresponding LEAD dataset. Our hardware system consists of a camera and a LiDAR. Though the FoV of the LiDAR is small, the range of detected depth is relatively larger. The detected depth is up to 500 meters, however, limited by the angle of incidence, generally, the effective range of depth is from 3 meters to 200 meters. For the camera, since we want to capture the image which contains the detail of the distant scene, we set the focal length of the camera to 12mm, the camera and LiDAR are calibrated.

% To evaluate the extended depth map, we use a scanner to generate the ground truth depth map. The ground truth data is accurate and dense enough for our evaluation. And the depth range of our ground truth is up to 400 meters which is much larger than KITTI 
% The LEAD dataset is the first dataset that focuses on the MEMS LiDAR and outdoor long-range distance estimation. Unlike KITTI, LEAD provides an accurate dense depth map for evaluation and MEMS LiDAR data. There are 16792 pairs of images and depth maps for training and 120 pairs for testing. We believe that our LEAD dataset can promote the development of the MEMS LiDAR.

% \begin{figure}[!h]
%     \centering
%     \includegraphics[width=\linewidth]{pic/ourdata.png}
%     % \vspace{-12pt}
%     \caption{ourdata}% The buildings and goal in our final depth result keeps decent continuity and integrity}
%     % \vspace{-12pt}
%     \label{fig:dataset}
% \end{figure}
% 

\paragraph{KITTI Eigen Split}

To evaluate the generalizability of our proposed method, we conduct experiments on the KITTI dataset~\cite{geiger2012we}. However, the original KITTI dataset provides only the ground truth depth but not partial depth maps with limit FoV like those from MEMS LiDAR in our setup. We resample a small region from the ground truth to simulate partial depth input. Note that we use the improved ground truth depth maps provided by KITTI in our experiments. Since the original ground truth depth maps are rendered from point clouds captured raw mechanical LiDAR with $360^\circ$ FoV, which is sparse and with many holes in rendered depth maps.% Similar to MonoDepth2~\cite{godard2019digging}, we use the KITTI Eigen Split~\cite{eigen2015predicting} to perform our evaluations. 

\begin{figure}
    \centering
    % \vspace{-8pt}
    \includegraphics[width=\linewidth]{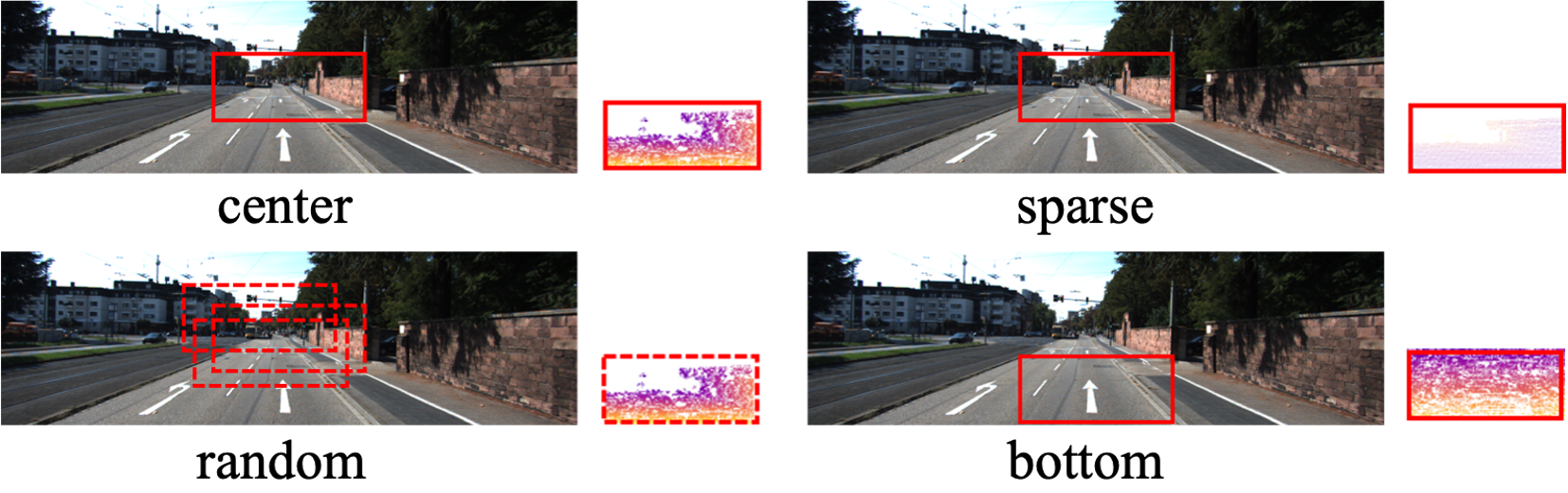}
    \caption{Different resampling modes in KITTI dataset.}% The buildings and goal in our final depth result keeps decent continuity and integrity}
    % \vspace{-8pt}
    \label{fig:cropping_modes}
\end{figure}

Regarding resampling from the ground truth, we provide 4 modes to simulate different MEMS LiDARs with different calibration. \figref{fig:cropping_modes} shows the 4 resampling modes: center cropping (center), downsampling on center cropping (sparse),  random cropping at the center region (random), and bottom cropping (bottom). Center cropping is the straightforward mode to simulate small FoV LiDAR located in the center of wide FoV camera. Sparse sampling mode simulates the MEMS can only provide very sparse points in small FoV. Bottom cropping simulates MEMS LiDAR yielding points in near and narrow space w.r.t the target view. In this setup, both the FoV and depth range need to be extended. Random cropping is a challenge setup, where the valid area of MEMS LiDAR locates on our target view randomly.

\begin{figure}
    \centering
    % \vspace{-12pt}
    \includegraphics[width=\linewidth]{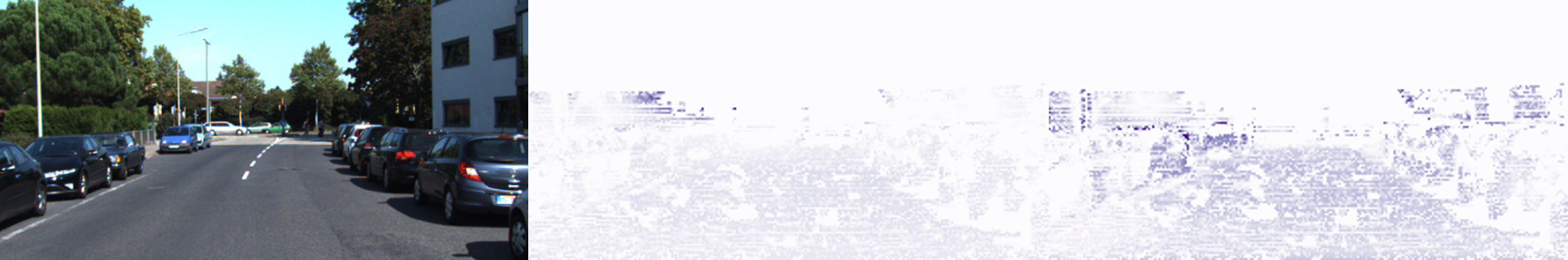}
    
    \caption{Qualitative results on multi-stage propagation. \textbf{LEFT}: the input RGB image. \textbf{MIDDLE}: depth map error after one stage propagation (darker color indicates more accurate depth). \textbf{RIGHT}: depth map error after all propagation stages.}
    % \vspace{-12pt}
    \label{fig:propagation qualitative}
\end{figure}

\begin{figure}
    \centering
    \includegraphics[width=\linewidth]{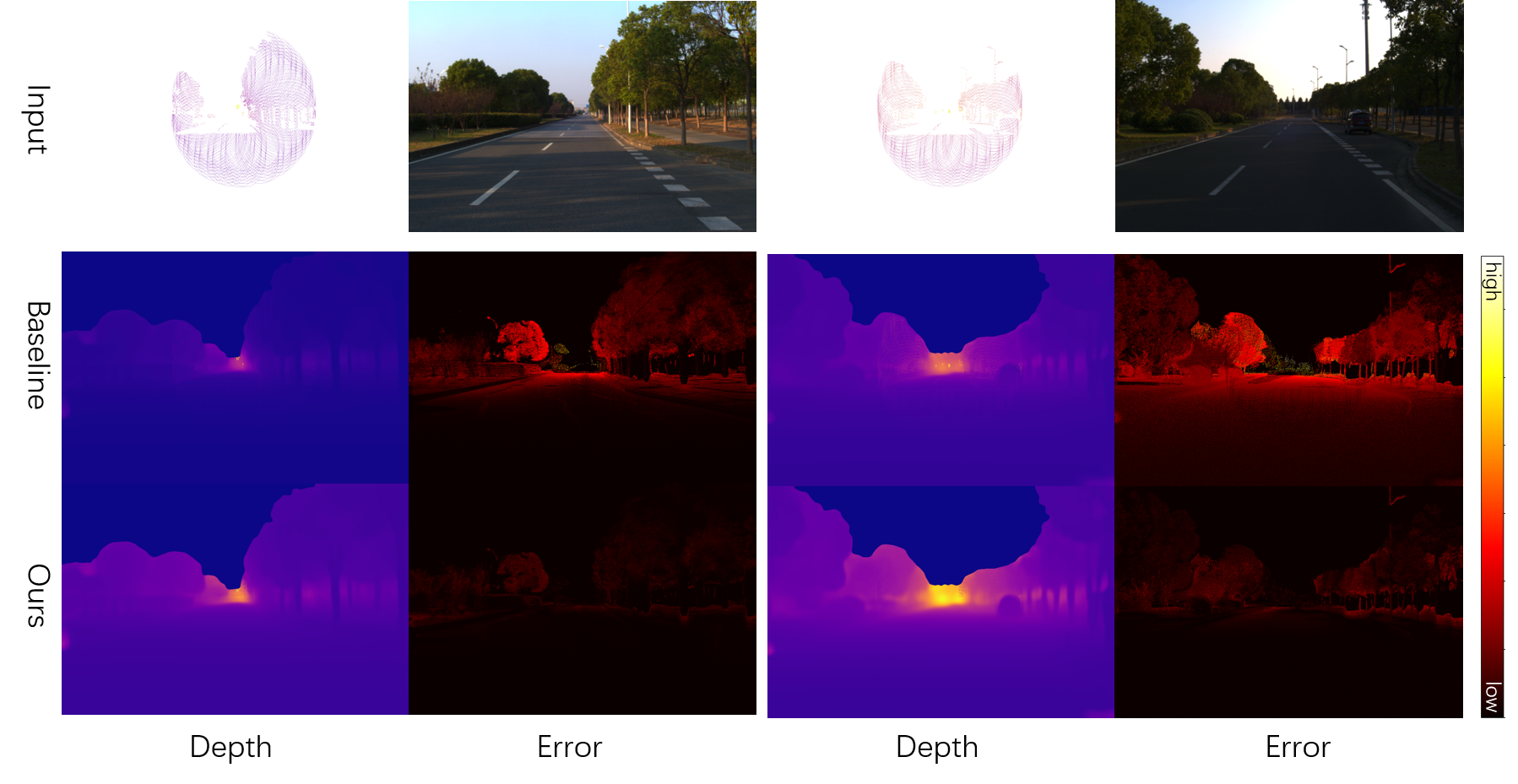}
    % \vspace{-12pt}
    \caption{Qualitative results on LEAD dataset.}% The buildings and goal in our final depth result keeps decent continuity and integrity}
    % \vspace{-12pt}
    \label{fig:lead qualitative}
\end{figure}

\begin{figure}
    \centering
    \includegraphics[width=\linewidth]{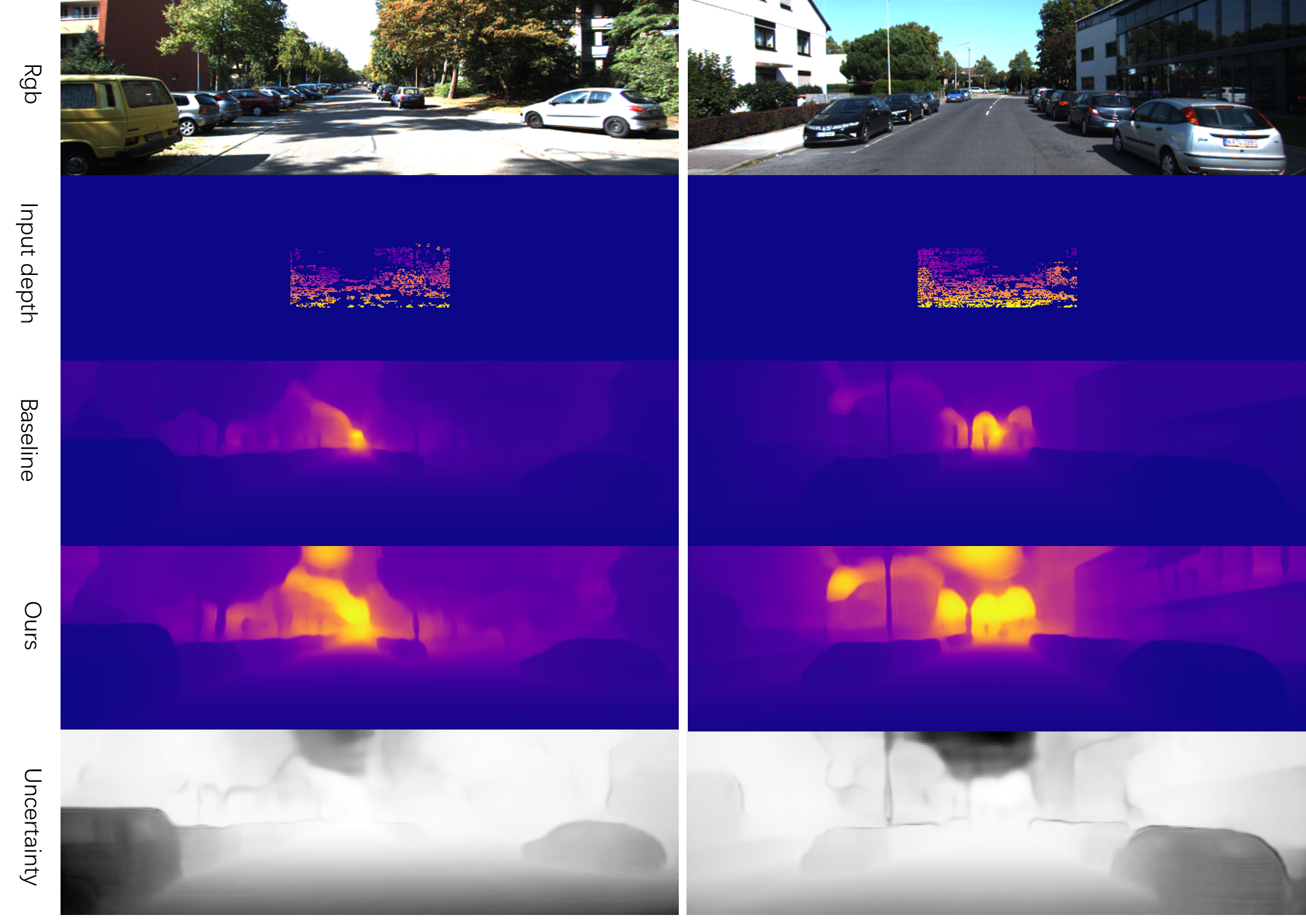}
    % \vspace{-12pt}
    \caption{Qualitative results on KITTI dataset.}% The buildings and goal in our final depth result keeps decent continuity and integrity}
    % \vspace{-12pt}
    \label{fig:kitti qualitative}
\end{figure}

\subsection{Qualitative Results}
\paragraph{Multi-stage propagation}
We illustrate the effectiveness of the multi-stage propagation mechanism. In particular, 
\figref{fig:propagation qualitative} shows that the depth map error reduces after all propagation stages (RIGHT) as compared to the depth map error after a single propagation stage (MIDDLE). 

\paragraph{Results on LEAD dataset}
\figref{fig:lead qualitative} shows the qualitative results on our LEAD dataset. The second row shows the depth map and its the error map for the STN module. The third row shows the results for our full pipeline.

\paragraph{Results on KITTI dataset}
\figref{fig:kitti qualitative} shows the qualitative results on the KITTI dataset. Since our network is based on MonoDepth2~\cite{godard2019digging}, we use it as our baseline method. We train the MonoDepth2 supervised by the input narrow FoV depth map. MonoDepth2 suffers from the problem of sparse and uneven depth distribution. The input narrow FoV depth map is not able to correct the scale even after using the ratio of the median to correct the monocular depth scale. By contrast, our propagation and distribution output modules are designed to solve this problem, resulting more accurate wide range of depth maps.

\begin{table*}[t]
    \begin{center}
        \begin{tabular}{ccccccccc}
        
        \hline
        \multicolumn{2}{c}{}    & \multicolumn{4}{c}{lower is better}                                                       & \multicolumn{3}{c}{higher is better}                                              \\ \cline{3-9} 
        \multicolumn{1}{c}{\multirow{-2}{*}{Method}} &\multicolumn{1}{c}{\multirow{-2}{*}{SC}} &\cellcolor[HTML]{9B9B9B}Abs Rel& \cellcolor[HTML]{9B9B9B}Sq Rel& \cellcolor[HTML]{9B9B9B}RMSE & \cellcolor[HTML]{9B9B9B}RMSE log &\cellcolor[HTML]{9999CC}$\delta_{1.25}$ & \cellcolor[HTML]{9999CC}$\delta_{1.25^2}$ & \cellcolor[HTML]{9999CC}$\delta_{1.25^3}$ \\ 
        \hline
        EPC++\cite{luo2018every}&M&0.120&0.789&4.755&0.177&0.856&0.961&0.987\\
        Geonet\cite{yin2018geonet}&M&0.132&0.994&5.240&0.193&0.833&0.953&0.985\\
        MonoDepth2\cite{godard2019digging}&M&0.090&0.545&3.942&0.137&0.914&0.983&0.995\\
        \hline
        EPC++&F&0.153&0.998&5.080&0.204&0.805&0.945&0.982\\
        Geonet&F&0.202&1.521&5.829&0.244&0.707&0.913&0.970\\
        MonoDepth2&F&0.109&0.623&4.136&0.154&0.873&0.977&0.994\\
        \hline
        MonoDepth2&P&0.104&0.560&3.898&0.48&0.889&0.980&0.995\\
        \textbf{Ours (res18)}&P&0.101&0.456&3.475&0.140&0.909&\textbf{0.984}&\textbf{0.996}\\
        \textbf{Ours (res50)}&P&\textbf{0.090}&\textbf{0.424}&\textbf{3.419}&\textbf{0.133}&\textbf{0.916}&\textbf{0.984}&\textbf{0.996}\\
        \hline
        \end{tabular}
        \caption{Quantitative results on KITTI Eigen Split with improved ground truth. ``SC'' means different scale correction methods: ``M'' means using different scales for each individual test case. The scale is the ratio between the median depth of the ground truth depth map and the median depth of the resulting depth map. ``F'' means using a fixed scale that is the mean scale. ``P'' means using the scale estimated from the input partial depth.}
        \label{tab:kitti quantitative}
        \vspace{-12pt}
    \end{center}
\end{table*}

% \begin{table}[!h]
%     \begin{center}
%     \resizebox{\linewidth}{!}{  
%         \begin{tabular}{cccccccc}
        
%         \hline
%         \multicolumn{1}{c}{}                         & \multicolumn{4}{c}{lower is better}                                                       & \multicolumn{3}{c}{higher is better}                                              \\ \cline{2-8} 
%         \multicolumn{1}{c}{\multirow{-2}{*}{Method}} & \cellcolor[HTML]{9B9B9B}Abs Rel& \cellcolor[HTML]{9B9B9B}Sq Rel& \cellcolor[HTML]{9B9B9B}RMSE & \cellcolor[HTML]{9B9B9B}RMSE log &\cellcolor[HTML]{9999CC}$\delta_{1.25}$ & \cellcolor[HTML]{9999CC}$\delta_{1.25^2}$ & \cellcolor[HTML]{9999CC}$\delta_{1.25^3}$ \\ \hline
%                                                     basline&0.437&6.134&15.718&0.633&0.070&0.174&0.603                     \\
%                                                      ours&0.110&1.039&6.299&0.171&0.859&0.963&0.991 \\
%                                                      \hline
%         \end{tabular}
%     }
%     \caption{Quantitative results on LEAD dataset. The baseline is our teacher network.}
%     \end{center}
% \end{table}

\begin{table}[!h]
    \begin{center}
    \resizebox{\linewidth}{!}{  
        \begin{tabular}{cccccccc}
        
        \hline
        \multicolumn{1}{c}{}                         & \multicolumn{2}{c}{lower is better}                                                       & \multicolumn{3}{c}{higher is better}                                              \\ \cline{2-6} 
        \multicolumn{1}{c}{\multirow{-2}{*}{Method}} & \cellcolor[HTML]{9B9B9B}RMSE & \cellcolor[HTML]{9B9B9B}RMSE log &\cellcolor[HTML]{9999CC}$\delta_{1.25}$ & \cellcolor[HTML]{9999CC}$\delta_{1.25^2}$ & \cellcolor[HTML]{9999CC}$\delta_{1.25^3}$ \\ \hline
                                                    STN&15.718&0.633&0.070&0.174&0.603                     \\
                                                     \textbf{Ours}&\textbf{6.299}&\textbf{0.171}&\textbf{0.859}&\textbf{0.963}&\textbf{0.991} \\
                                                     \hline
                                            STN ($<80$)&  14.446  &   0.633  &   0.072  &   0.176  &   0.604                      \\
                                             \textbf{Ours ($<80$)}&   \textbf{6.053}  &   \textbf{0.172}  &   \textbf{0.858}  &   \textbf{0.963}  &   \textbf{0.991} \\ \hline
                                             STN ($>80$)&  22.687  &   0.330  &   0.720  &   0.743  &   0.848                     \\
                                             \textbf{Ours ($>80$)}&   \textbf{6.968}  &   \textbf{0.077}  &   \textbf{0.959}  &   \textbf{0.999}  &   \textbf{1.000} \\
                                             \hline
                                             
        \end{tabular}
    }
    \caption{Quantitative results on LEAD dataset.}
    
    \label{tab:lead quantitative}
    \end{center}
\end{table}

\subsection{Quantitative Results}
\paragraph{Results on LEAD dataset}
Tab. \ref{fig:lead qualitative} shows the quantitative results on our LEAD dataset. We compare results for the STN module and our full pipeline. Moreover, since our LEAD dataset contains ground truth depth over 80 meters that is the maximum ground truth depth provided by the KITTI dataset, we also show results using the ground truth depth less than 80 meters and larger than 80 meters respectively.

% Considering the target of our extender is to make the depth map wider and deeper, we also show 

\paragraph{Results on KITTI dataset}
Tab.~\ref{tab:kitti quantitative} shows the quantitative results on the KITTI Eigen Split~\cite{eigen2015predicting}. We compare our results with various monocular depth estimation methods. The first and second row blocks shows results for the unsupervised monocular depth estimation methods, which needs scale correction based on ground truth depth information. However, using depth information from ground truth is not a practical assumption. By contrast,
our method only needs partial depth maps. Hence, the third row block shows results for MonoDepth2 and our method using partial depth map for scale correction. The results show that our method outperforms the baseline methods.

Tab.~\ref{tab:resample} shows the quantitative results of our method on KITTI dataset under different resampling modes. The results do not differ much under different resampling modes, which indicates the generalizability of our method.

% The monocular depth estimation methods use the KITTI Eigen Split~\cite{eigen2015predicting} to evaluate the performance. Considering that we need the dense ground truth as input, we use the improved ground truth that is much more accurate than raw LiDAR depth. 

% Here we compare our results with the monocular depth estimation methods. The unsupervised mono depth estimation methods rely on a scale that can only be obtained from the ground truth data.

% \begin{equation}
%     scale = median(D_{GT})/median(D_{res}),
% \end{equation}    

% Without the ground truth, these methods cannot estimate accurate depth. The scale is not stable, so for each estimated depth map, these methods need to calculate the corresponding scale with the full ground truth depth map while evaluating. These methods can also use a fixed scale that is the mean value of all the scale value. 

% But actually, this kind of scale correction method is not feasible, because the full ground truth is not available and if we get the full ground truth, we do not have to estimate the depth at all.

% For our method, we do not need the whole ground truth to solve the scale problem. We rely on the partial depth which can be obtained in the outdoor environment. We also try to use the partial depth to correct the scale of the unsupervised mono depth algorithm and compare the result with ours. The table below shows that with small FoV ground truth, our results are much better.

% While testing MonoDepth2, we correct the scale with the partial depth. And we also show the result of MonoDepth2 supervised by partial depth. 

\paragraph{Ablation Study}

%We conduct an ablation study using the KIITI dataset.
Tab.~\ref{tab:kitti ablation} shows the ablation study results. We compare the performances among three methods: STN, STN+PPG, and STN+PPG+PDC. 
STN is the normal propagation without distribution generation. STN+PPG adds the distribution generation. STN+PPG+PDC is our entire pipeline.
It can be shown that the proposed PPG and PDC modules improve the performance effectively.

\begin{table}[h]
\begin{tabular}{cccccc}
\hline
\multicolumn{1}{c}{}                         & \multicolumn{2}{c}{lower is better}                                                       & \multicolumn{3}{c}{higher is better}                                              \\ \cline{2-6} 
\multicolumn{1}{c}{\multirow{-2}{*}{Method}} & \cellcolor[HTML]{9B9B9B}RMSE & \cellcolor[HTML]{9B9B9B}RMSE log &\cellcolor[HTML]{9999CC}$\delta_{1.25}$ & \cellcolor[HTML]{9999CC}$\delta_{1.25^2}$ & \cellcolor[HTML]{9999CC}$\delta_{1.25^3}$ \\ \hline
                                            Center&  3.475  &   0.140  &   0.909  &   0.984  &   0.996 \\ \hline
                                            Sparse&  3.558  &   0.140  &   0.909  &   0.983  &   0.995 \\ \hline   
                                            Random&   3.483  &   0.141  &   0.909  &   0.984  &   0.996 \\ \hline
                                            Bottom&   3.859  &   0.136  &  0.911  &   0.982  &   0.995 \\ \hline    
\end{tabular}
\caption{Results on different resampling modes.}

\label{tab:resample}
\end{table}

\begin{table}[!h]
    \begin{center}
        \resizebox{\linewidth}{!}{  
            \begin{tabular}{cccccccc}
            
            \hline
            \multicolumn{1}{c}{}                         & \multicolumn{4}{c}{lower is better}                                                       & \multicolumn{3}{c}{higher is better}                                              \\ \cline{2-8} 
            \multicolumn{1}{c}{\multirow{-2}{*}{Method}} & \cellcolor[HTML]{9B9B9B}Abs Rel& \cellcolor[HTML]{9B9B9B}Sq Rel& \cellcolor[HTML]{9B9B9B}RMSE & \cellcolor[HTML]{9B9B9B}RMSE log &\cellcolor[HTML]{9999CC}$\delta_{1.25}$ & \cellcolor[HTML]{9999CC}$\delta_{1.25^2}$ & \cellcolor[HTML]{9999CC}$\delta_{1.25^3}$ \\ \hline
            STN&0.103&0.504&3.713&0.148&0.886&0.979&0.995                     \\
            STN+PPG&0.106&0.486&3.635&0.144&0.904&0.983&0.996 \\
            STN+PPG+PDC&0.101&0.456&3.475&0.140&0.909&0.984&0.996\\
                                                         \hline
            \end{tabular}
        }
        \caption{Ablation study on KITTI dataset.}
        
        \label{tab:kitti ablation}
    \end{center}
\end{table}

\section{Conclusion}

We propose a novel self-supervised learning based method to use the small field of view solid-state MEMS LiDAR to estimate a deeper and wider depth map, a new hardware system, and a corresponding dataset for this kind of LiDAR. The experiment shows that our new setup and algorithm make full use of the depth generated by the MEMS LiDAR and generate a deeper and wider depth map. We believe that our setup and method can promote the wide application of the low cost and miniaturized solid-state MEMS LiDAR in the field of autonomous driving.

%-------------------------------------------------------------------------

%-------------------------------------------------------------------------

{\small

%\bibliographystyle{ieee_fullname}
%\bibliography{egbib}
}

\end{document}